\documentclass{article}
\usepackage{spconf,amsmath,graphicx,hyperref}

\usepackage{amsmath,amssymb,amsfonts}
\usepackage{bm}
\usepackage{enumitem}

\DeclareMathOperator{\Logm}{logm}        
\DeclareMathOperator{\softmax}{softmax}
\DeclareMathOperator{\diag}{diag}
\DeclareMathOperator{\pool}{pool}

\title{NeuroBRIDGE: Behavior-Conditioned Koopman Dynamics with Riemannian Alignment for Early Substance Use Initiation Prediction from Longitudinal Functional Connectome}
\name{Badhan Mazumder$^{\star \dagger}$ \qquad Sir-Lord Wiafe$^{\star \dagger}$ \qquad Vince D. Calhoun$^{\star \dagger}$ \qquad Dong Hye Ye$^{\star \dagger}$}
  
  \address{$^{\star}$ Department of Computer Science, Georgia State University\\
      $^{\dagger}$Tri-Institutional Center for Translational Research in Neuroimaging and Data Science (TReNDS),\\ Georgia State University, Georgia Institute of Technology, and Emory University}
\makeatletter
\let\oldthebibliography\thebibliography
\def\thebibliography#1{%
  \oldthebibliography{#1}%
  \setlength{\itemsep}{0pt}%
  \setlength{\parskip}{0pt}%
  \setlength{\parsep}{0pt}%
}
\makeatother
\begin{document}
%
\maketitle
\begin{abstract}

Early identification of adolescents at risk for substance use initiation (SUI) is vital yet difficult, as most predictors treat connectivity as static or cross-sectional and miss how brain networks change over time and with behavior. We proposed \textit{NeuroBRIDGE} (\textbf{B}ehavior conditioned \textbf{RI}emannian Koopman \textbf{D}ynamics on lon\textbf{G}itudinal conn\textbf{E}ctomes), a novel graph neural network-based framework that aligns longitudinal functional connectome in a Riemannian tangent space and couples dual-time attention with behavioral-conditioned Koopman dynamics to capture temporal change. Evaluated on ABCD, \textit{NeuroBRIDGE} improved future SUI prediction over relevant baselines while offering interpretable insights into neural pathways, refining our understanding of neurodevelopmental risk and informing targeted prevention.

\end{abstract}
\begin{keywords}
Substance Use Initiation, Functional Connectome,  Riemannian Manifold, Graph Neural Network, Neural Koopman Operator, Longitudinal Analysis
\end{keywords}
\vspace{-10pt}
\section{Introduction}
 \vspace{-10pt}
Adolescence is a period of rapid remodeling in reward, control, and social circuits \cite{I_1}. It is also when many youths become susceptible to substance use, so early risk detection is crucial. Functional network connectivity (FNC) from resting-state functional magnetic resonance imaging (rs-fMRI) offers a systems-level view how these circuits evolve and can reveal imbalances before they appear behaviorally \cite{My_3,My_4}. On the behavioral side, caregiver-reported Child Behavior Checklist (CBCL) scores, especially measures of externalizing tendencies, provide a practical read on psychosocial risk that tracks with day-to-day functioning. Bringing these two sources together let us capture both the neurobiological trajectory and the behavioral context of risk, creating a clinically meaningful pathway for earlier, better-targeted prevention.

Despite recent progress in deep learning based computational neuroimaging \cite{My_3,My_4,My_1,My_2}, current methods face three persistent roadblocks. First, a reliance on static or cross-sectional modeling. Architectures that operate on a single connectome (e.g., CNN- \cite{BrainNetCNN} or transformer-style encoders \cite{BNT}) can learn topology, yet they miss the within-person trajectories that carry developmental patterns across visits. Second, temporal models that ignore geometry. Graph neural network (GNN) based dynamic approaches \cite{Evolvegcn,RGBM} track change over time, but they typically treat connectivity as Euclidean data. FNC matrices live on a curved (Riemannian) space; flattening them can distort distances, amplify noise, and destabilize training. Third, limited use of behavioral covariates. Many pipelines \cite{BrainNetCNN,BNT,RGBM} employed FNC alone, leaving out psychosocial markers like CBCL scores that are known to modulate both brain maturation and early use propensity. These gaps collectively blunt predictive power and make it harder to extract clinically meaningful insight.

To address these challenges, we proposed \textit{NeuroBRIDGE}, a geometry-aware GNN based framework that unifies longitudinal connectome modeling with behavioral conditioning. \textit{NeuroBRIDGE} anchors each subject’s baseline and follow-up FNCs in a Riemannian tangent space, generates multi-scale node embeddings through heat-kernel tokens, and leverages a dual-time fusion module to jointly learn within-visit and cross-visit dependencies. A neural Koopman operator, conditioned on CBCL scores, further models spectral dynamics of latent evolution while embedding behavioral context.
\begin{figure*}[htb]
  \centering
  \centerline{\includegraphics[width=\linewidth]{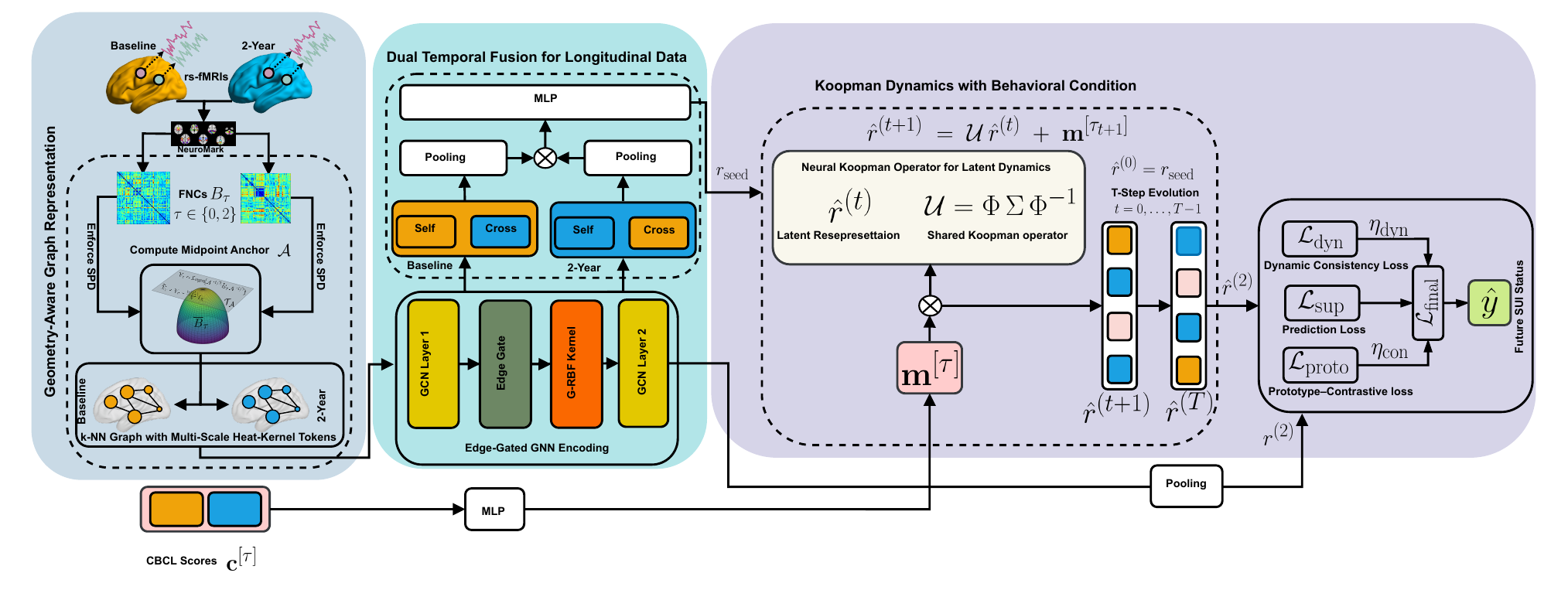}}
  \vspace{-10pt}
\caption{\textbf{\textit{NeuroBRIDGE} overview:} Longitudinal rs-fMRIs (Baseline and 2-Year) were processed to generate FNCs $(53\times 53)$ via $NeuroMark$, then SPD-regularized and aligned in a shared Riemannian tangent space to form geometry-aware \(k\)NN graphs with multi-scale heat-kernel tokens as node features followed by an edge-gated GNN encoding visit-specific embeddings, which are fused by dual-time self/cross attention into a subject-wise vector. A spectrally bounded, CBCL-conditioned Koopman module then models longitudinal change, and employed the predicted–observed year-2 latent to estimate 4-year SUI status, jointly optimized by three different loss terms in an end-to-end manner.}
  \vspace{-13pt}
\label{fig:fig1}
\end{figure*}

The key contributions of this work are threefold:
\begin{enumerate}
    \item  We employed a midpoint-anchored tangent-space embedding with multi-scale tokens that respects the manifold structure of FNC.
     \vspace{-10pt}
    \item We developed a temporal fusion layer that captured both intra- and inter-timepoint dependencies to better characterize developmental change.
     \vspace{-10pt}
    \item We introduced a spectral Koopman module conditioned on CBCL scores to jointly model FNC evolution and psychosocial risk, enabling improved SUI prediction.
     \vspace{-10pt}
\end{enumerate}

\vspace{-8pt}
\section{Methodology}
 \vspace{-8pt}
For each subject, let \(\{B_{\tau}\}_{\tau\in\{0,2\}}\subset\mathbb{R}^{N\times N}\) denote baseline and 2-year FNC (\(N=53\) brain networks), and let \(\{\mathbf{c}^{[\tau]}\}_{\tau\in\{0,2\}}\subset\mathbb{R}^{F}\) be the corresponding CBCL features (\(F=6\)). The prediction target is \(y\in\{0,1\}\) indicating SUI status at the 4-year follow-up (\(1=\) SUI; \(0=\) non-SUI). The goal is to learn $ \{B_{\tau},\,\mathbf{c}^{[\tau]}\}_{\tau\in\{0,2\}}\ \longmapsto\ \hat y\in[0,1]$ where \(\hat{y}\) estimates the probability of future SUI, by jointly modeling geometry-aware longitudinal FNC representations with behavioral scores as depicted in Fig. \ref{fig:fig1}.
\vspace{-15pt}
\subsection{Geometry-Aware Graph Representation}
\vspace{-5pt}
\subsubsection{Riemannian Tangent-Space Embedding }
\vspace{-5pt}
FNCs naturally lie on the Riemannian manifold of symmetric positive-definite (SPD) matrices; treating them in Euclidean space distorts geodesic structure and undermines baseline–follow-up comparability. We therefore (i) enforced SPD, (ii) used the affine-invariant Riemannian geometry to map both sessions into a single subject-anchored tangent space at the geodesic midpoint (so changes can be measured in the same coordinates), and (iii) removed the isotropic mode to reduce scanner drift while preserving anisotropy.


At first SPD was enforced on $B_{\tau}$ via a small ridge: $\overline{B}_{\tau}=\tfrac{1}{2}(B_{\tau}+B_{\tau}^{\top})+\epsilon I_{N},\ \epsilon>0,\ \tau\in\{0,2\}$. Then the affine-invariant \cite{M_3} midpoint (anchor) was computed as:
\begin{equation}\label{eq:anchor}
\mathcal{A}
= B_{0}^{1/2}\!\big(B_{0}^{-1/2}B_{2}B_{0}^{-1/2}\big)^{1/2}\!B_{0}^{1/2}
\end{equation}
and  each session was mapped to the shared tangent space $\mathcal{T}_{\mathcal{A}}$ via the principal matrix logarithm $\Logm(\cdot)$ \cite{M_2}:
\begin{equation}\label{eq:logmap}
Y_{\tau}
=\Logm\!\big(\mathcal{A}^{-1/2}\,\overline{B}_{\tau}\,\mathcal{A}^{-1/2}\big)
\end{equation}
After that, the isotropic mode was removed and rows was considered as node features:
\(
\widetilde{Y}_{\tau}=Y_{\tau}-\tfrac{\mathrm{tr}(Y_{\tau})}{N}I_{N},\ 
y^{(\tau)}_{i}=\widetilde{Y}_{\tau}(i,:),\ i=1,\ldots,N.
\)
Here $\Logm(\cdot)$ is the principal matrix logarithm, $I_{N}$ defines identity, and $\mathrm{tr}(\cdot)$ is the trace. 


\vspace{-10pt}

\subsubsection{Multi-Scale Heat-Kernel Tokens \& Edge-Gated GNN Encoding}
\vspace{-5pt}
\noindent\textit{Heat-Kernel Tokens.} Before GNN encoding, we built heat-kernel (H-K) tokens to capture local-to-global structure because the graph heat kernel \(e^{-\theta\mathcal{J}_\tau}\) is a stable low-pass whose scale \(\theta\) smoothly controls context \cite{Diffusion_kernels}. We constructed a $k$NN graph on $\{y^{(\tau)}_{i}\}$ with adjacency $G_{\tau}$ and edge set $\mathcal{E}_{\tau}$. Let
\(
\Gamma_{\tau}=\diag(G_{\tau}\mathbf{1}),\ 
\widehat{G}_{\tau}=\Gamma_{\tau}^{-1/2}G_{\tau}\Gamma_{\tau}^{-1/2},\ 
\mathcal{J}_{\tau}=I-\widehat{G}_{\tau},
\)
where $\mathbf{1}$ is the all-ones vector.
For learnable scales \(\{\theta_s\}_{s=1}^{S}\), H-K tokens are  $\Psi^{(\tau)}_{0}=\widetilde{Y}_\tau$,
$\Psi^{(\tau)}_{s}=\exp(-\theta_s \mathcal{J}_\tau)\,\widetilde{Y}_\tau$\ which were then concatenated and projected row-wise to obtain compact node features:
\begin{equation}
\breve{Y}_\tau=\mathrm{MLP}_{\text{row}}\!\Big(\big[\Psi^{(\tau)}_{0}\,\Vert\,\cdots\,\Vert\,\Psi^{(\tau)}_{S}\big]\Big)\in\mathbb{R}^{N\times D_0}.
\end{equation}
where \(\exp(\cdot)\) is the matrix exponential and \(\mathrm{MLP}_{\text{row}}\) is a lightweight projector applied per row.

\noindent\textit{Geometry-guided gating \& GNN encoding.}
To favor edges between geometrically close nodes in the shared tangent space and suppress spurious long-range connections, we modulated each candidate edge with a tangent–space Gaussian RBF kernel \cite{GRBF_kernel}
\(\Pi^{(\tau)}_{ij}=\exp\!\big(-\|y^{(\tau)}_{i}-y^{(\tau)}_{j}\|_{2}^{2}/(2\,\omega_{\tau}^{2})\big)\) for \((i,j)\!\in\!\mathcal{E}_{\tau}\) (bandwidth \(\omega_{\tau}\) from neighbor distances). Firstly, we encoded with a gated two-layer
graph convolution network (GCN) \cite{MM_0} where node states are first computed as \(R^{(\tau)}_{1}=\mathrm{GCN}_{1}(\breve{Y}_{\tau},\widehat{G}_{\tau})\). Then the edge gate $g^{(\tau)}_{ij}=\sigma\!\Big(\mathrm{MLP}_{e}\big([R^{(\tau)}_{1,i};\,R^{(\tau)}_{1,j}]\big)\Big)$ was predicted and the RBF kernel was applied as: $\tilde g^{(\tau)}_{ij}=g^{(\tau)}_{ij}\Pi^{(\tau)}_{ij}$ yielding a reweighted adjacency $G^{\star}_{\tau}(i,j)=\tilde g^{(\tau)}_{ij}\,G_{\tau}(i,j)$, and 
$\widehat{G}^{\star}_{\tau}=D^{\star-1/2}_{\tau}G^{\star}_{\tau}D^{\star-1/2}_{\tau}$ as its  normalized form  where  $D^{\star}_{\tau}=\diag(G^{\star}_{\tau}\mathbf{1})$, to produce the final node embeddings as:
\begin{equation}\label{eq:gcn2}
R^{(\tau)}=\mathrm{GCN}_{2}\!\big(R^{(\tau)}_{1},\,\widehat{G}^{\star}_{\tau}\big)
\end{equation}
where \(\sigma(\cdot)\) is the logistic function and \(\mathrm{MLP}_{e}\) presents a small projector.
\vspace{-10pt}
\subsection{Dual Temporal Fusion for Longitudinal Data}

Instead of simple concatenation, we fused node embeddings with both self + cross-attention so each visit queries itself and the other in a shared space, emphasizing baseline→year-2 reorganization and cross-time dependencies that better predict SUI than either scan alone.

Using shared projections \(W_Q,W_K,W_V\!\in\!\mathbb{R}^{D\times D}\), we formed
\(Q_{\tau}=R^{(\tau)}W_Q,\ K_{\tau}=R^{(\tau)}W_K,\ V_{\tau}=R^{(\tau)}W_V\) (row-wise softmax for attention), and a scalar gate \(\pi=\sigma(\upsilon)\in(0,1)\).
Let \(\bar\tau=2-\tau\) denote the other timepoint. The fused node representation is
\begin{equation}\label{eq:dual_fusion}
\begin{aligned}
\widetilde R^{(\tau)}
&= \pi\,\softmax\!\Big(\tfrac{Q_{\tau}K_{\tau}^{\top}}{\sqrt{D}}\Big)V_{\tau} \\
&\quad + (1-\pi)\,\softmax\!\Big(\tfrac{Q_{\tau}K_{\bar\tau}^{\top}}{\sqrt{D}}\Big)V_{\bar\tau}
\end{aligned}
\end{equation}

We obtained subject-wise vectors by pooling \(\tilde r^{(\tau)}=\pool(\widetilde R^{(\tau)})\in\mathbb{R}^{D}\). A small MLP was used to produce mixture weights 
\(\bm{\delta}=\softmax(\mathrm{MLP}([\tilde r^{(0)};\tilde r^{(2)}]))\in\mathbb{R}^{2}\) (\(\delta_0{+}\delta_2{=}1\)), and the fused seed for dynamics can be formulated as  
\(r_{\mathrm{seed}}=\delta_0\,\tilde r^{(0)}+\delta_2\,\tilde r^{(2)}\).
\vspace{-10pt}
\subsection{Koopman Dynamics with Behavioral Condition}

\subsubsection{Neural Koopman Operator for Latent Dynamics}
Inspired by the Koopman operator's \cite{MM_2} viewpoint  on nonlinear dynamical systems, we aimed to capture how the fused representation evolves from baseline to 2-year in a stable manner, while conditioning the evolution on CBCL scores to reflect psychosocial influences on brain-state change.

We parameterized a shared linear operator in the \(D\)-dimensional latent space (with \(R^{(\tau)}\!\in\!\mathbb{R}^{N\times D}\), \(r\!\in\!\mathbb{R}^{D}\)) and injected CBCL scores at each visit: \(\mathcal{U}=\Phi\,\Sigma\,\Phi^{-1}\) with \(\Sigma=\diag(\tanh(\chi_1),\ldots,\tanh(\chi_D))\), \(\Phi\!\in\!\mathbb{R}^{D\times D}\), \(\chi\!\in\!\mathbb{R}^{D}\); CBCL offsets are \(\mathbf{m}^{[\tau]}=\mathrm{MLP}_{c}(\mathbf{c}^{[\tau]})\in\mathbb{R}^{D}\) (with \(\mathrm{MLP}_{c}:\mathbb{R}^{F}\!\to\!\mathbb{R}^{D}\)). The \(\tanh\) bounding of \(\Sigma\) constrains the spectrum to \((-1,1)\) for stable evolution. The $T$ step rollout from the fused representation can be formed as follows: 
\begin{equation}
\hat r^{(t+1)} \;=\; \mathcal{U}\,\hat r^{(t)} \;+\; \mathbf{m}^{[\tau_{t+1}]},\quad t=0,\ldots,T\!-\!1 .
\end{equation}
where $\hat r^{(0)}=r_{\mathrm{seed}}$ and  \(\{\tau_1,\ldots,\tau_T\}\) are the visit indices providing controls (in our study \(T{=}2\): \(\tau_1{=}0\), \(\tau_2{=}2\)), so
\(\hat r^{(1)}=\mathcal{U}r_{\mathrm{seed}}+\mathbf{m}^{[0]}\) and
\(\hat r^{(2)}=\mathcal{U}\hat r^{(1)}+\mathbf{m}^{[2]}\).

\vspace{-10pt}
\subsubsection{Joint Training Objective}
Let \(r^{(2)}\) be the pre-fusion pooled year-2 vector after Equ.\ref{eq:gcn2}. We formed a deviation \(\Xi=r^{(2)}-\hat r^{(2)}\), concatenated \([\hat r^{(2)};\Xi]\), projected to \(\mathbf{g}_{\mathrm{cls}}=\varphi([\hat r^{(2)};\Xi])\), obtained \(\ell_{\mathrm{logit}}=\mathrm{MLP}_{\mathrm{cls}}(\mathbf{g}_{\mathrm{cls}})\), and predicted \(\hat y=\sigma(\ell_{\mathrm{logit}})\). The supervised loss uses class-weighted BCE with positive-class weight \(\beta_{+}\): \(\mathcal{L}_{\mathrm{sup}}=\mathrm{BCE}_{\beta_{+}}(\ell_{\mathrm{logit}},y)\). To align the rollout with the observed year-2 state, we included a consistency loss term \(\mathcal{L}_{\mathrm{dyn}}=\|\hat r^{(2)}-r^{(2)}\|_2^2\). Yet matching the year-2 state does not ensure class separation—especially under class imbalance; so we employed a prototype–contrastive (minority-aware) term \cite{prototype_loss} as a regularizer. We normalized features \(\widehat{\mathbf{g}}=\mathbf{g}_{\mathrm{cls}}/\|\mathbf{g}_{\mathrm{cls}}\|\) and kept unit-norm class prototypes \(\nu_c\) (\(c\!\in\!\{0,1\}\)) via exponential moving average (EMA): \(\nu_c\!\leftarrow\!\mathrm{normalize}\!\big(\zeta\,\nu_c+(1-\zeta)\,\overline{\widehat{\mathbf{g}}}_c\big)\), where \(\overline{\widehat{\mathbf{g}}}_c\) is the batch mean and \(\zeta\!\in\![0,1)\). Similarity uses temperature-scaled cosine logits \(\kappa_c=\langle\widehat{\mathbf{g}},\nu_c\rangle/\varsigma_c\) (class temperatures \(\varsigma_{c}\!>\!0\)). For label \(y\) and complement \(\bar y\), by setting \(\kappa_y=\kappa_{c=y}\), \(\kappa_{\bar y}=\kappa_{c=\bar y}\), the prototype–contrastive loss can be defined as $\mathcal{L}_{\mathrm{proto}}
= -\log\frac{\exp(\kappa_y)}{\exp(\kappa_y)+\exp(\kappa_{\bar y})}$.
This attracts features to the correct prototype and repels them from the other, improving separation under class imbalance.

The final objective combines all three terms as follows:
\begin{align}
\mathcal{L}_{\mathrm{final}}
= \mathcal{L}_{\mathrm{sup}}
+ \eta_{\mathrm{dyn}}\,\mathcal{L}_{\mathrm{dyn}}
+ \eta_{\mathrm{con}}\,\mathcal{L}_{\mathrm{proto}}.
\end{align}
where $\eta_{\mathrm{dyn}}$ and $\eta_{\mathrm{con}}$ weight the dynamics-consistency and prototype–contrastive terms, respectively.

\vspace{-10pt}
\section{Results and Discussion}
\vspace{-10pt}
\subsection{Dataset and Pre-processing}
\vspace{-5pt}
We utilized the Adolescent Brain Cognitive Development (ABCD) study \cite{D_1}; a multi-site U.S. longitudinal cohort across 21 sites. Baseline (age 9–10) and 2-year follow-up (age 11–12) data were utilized in our work for 7168 participants with resting-state fMRI and CBCL scores; 4-year (age 13–14) follow-up labels identified 220 youths with SUI.

Standard steps (slice-timing correction, motion realignment, spatial normalization, smoothing) were applied to rs-fMRI. Using the $NeuroMark$ \cite{D_2} ICA pipeline, we extracted 53 intrinsic connectivity networks (ICNs); their time series were bandpass filtered (0.01–0.15\,Hz) and z-scored. For each visit \(\tau\!\in\!\{0,2\}\), FNC matrices \(B_{\tau}\!\in\!\mathbb{R}^{53\times 53}\) were computed via normalized Dynamic Time Warping (nDTW) \cite{D_3} between ICN time courses. The nDTW window length followed the low-frequency cutoff, yielding 110 samples given TR\(=0.8\)s. 

\vspace{-10pt}
\subsection{Experimental Settings}
\vspace{-5pt}
We selected hyperparameters via 5-fold CV (80:20 splits). The final configuration used Adam (lr \(1{\times}10^{-4}\), weight decay \(5{\times}10^{-4}\)), batch size \(64\), dropout \(0.40\), and \(60\) epochs; \(k{=}5\) for \(k\)NN; H-K token steps \(S{=}2\); dimensions \(D_0{=}96\) (projector) and \(D{=}96\) (GNN hidden/latent). We enforced SPD with \(\epsilon{=}10^{-3}\). Dual-time attention used a fusion gate initialized at \(\pi{=}0.50\). The loss combined class-weighted BCE with dynamics and prototype terms (\(\eta_{\mathrm{dyn}}{=}0.25\), \(\eta_{\mathrm{con}}{=}0.10\)); the prototype head used EMA momentum \(\zeta{=}0.97\) and class temperatures \(\varsigma_{+}{=}0.08\) (SUI) and \(\varsigma_{-}{=}0.12\) (non-SUI).

\vspace{-10pt}
\subsection{Quantitative Evaluation}
\vspace{-5pt}

\textit{Baselines.} We benchmarked against 5 representative families: a manifold network (SPDNet \cite{SPDNet}), a CNN based (BrainNetCNN \cite{BrainNetCNN}), a transformer style model (BNT \cite{BNT}), and two temporal GNNs (EvolveGCN \cite{Evolvegcn}, RBGM \cite{RGBM}). Table~\ref{Table_1} demonstrates that \textit{NeuroBRIDGE} attained the best accuracy/sensitivity/specificity ($84.71/86.36/84.66\%$). Relative to SPDNet and BrainNetCNN which treated each connectome largely in isolation our subject wise midpoint-anchored Riemannian alignment plus dual-time fusion better captured cross-visit change, yielding $~7–8$ percentage points (pp) gain in accuracy and markedly higher sensitivity. Against BNT, which employed global attention but ignored the SPD geometry and longitudinal coupling, we gained $+5.7$ pp accuracy and $+11.9$ pp sensitivity, indicating that geometry-aware alignment and explicit self/cross attention matter for early-risk detection. Temporal GNNs (EvolveGCN, RBGM) tracked dynamics but lack our tangent-space anchoring and behavior-conditioned evolution. In contrast, \textit{NeuroBRIDGE} improved accuracy by $~9–10$ pp and raises sensitivity by $~13–19$ pp, highlighting the benefit of coupling longitudinal brain changes with CBCL-conditioned latent dynamics. 


\textit{Ablations.} As reported in Table~\ref{Table_2} we structured our ablation study around four primary research questions (RQs). \emph{RQ1--Riemannian alignment:} replacing the manifold pipeline with Euclidean dropped accuracy to \(75.27\%\) ($-9.44$ pp); midpoint anchoring and trace-free deflation offered smaller but consistent gains. \emph{RQ2--local-global tokens \& edge kernel:} removing H-K tokens ($-4.63$ pp), the edge gate ($-2.00$ pp), or the Gaussian RBF kernel ($-0.92$ pp) degraded performance. \emph{RQ3--cross-time fusion \& koopman dynamics with behavior:} without cross-time attention fusion (concatenation only) yielded \(77.92\%\) ($-6.79$ pp); without Koopman \(79.69\%\) ($-5.02$ pp) and omitting CBCL reduced accuracy to \(82.13\%\) ($-2.58$ pp). \emph{RQ4--loss terms \& class shaping:} without the dynamics term \(83.08\%\) ($-1.63$ pp) and removing the prototype--contrastive term reduced sensitivity to \(71.36\%\) ($-15.0$ pp). 

\begin{table}[t]
\caption{Performance comparison against sate-of-the-art (SOTA) baselines [Unit: \%] (Mean $\pm$ Standard Deviation).}
\vspace{-15pt}
\begin{center}
\resizebox{\columnwidth}{!}{\begin{tabular}{|c|c|c|c|} 
\hline
\textbf{Method} & \textbf{Accuracy} & \textbf{Sensitivity} & \textbf{Specificity} \\
\hline
SPDNet \cite{SPDNet}    & 77.57 $\pm$ 0.0854 & 68.18 $\pm$ 0.1122 & 77.86 $\pm$ 0.0811 \\ \hline
BrainNetCNN \cite{BrainNetCNN} & 76.53 $\pm$ 0.0932 & 68.64 $\pm$ 0.1213 & 76.78 $\pm$ 0.0866 \\ \hline
BNT  \cite{BNT}        & 79.02 $\pm$ 0.0794 & 74.42 $\pm$ 0.0976 & 79.16 $\pm$ 0.0754 \\ \hline
EvolveGCN \cite{Evolvegcn}    & 74.68 $\pm$ 0.0989 & 67.27 $\pm$ 0.1281 & 74.91 $\pm$ 0.0891 \\ \hline
RBGM \cite{RGBM}        & 75.73 $\pm$ 0.0912 & 72.82 $\pm$ 0.1056 & 75.85 $\pm$ 0.0844 \\ \hline
\textbf{\textit{NeuroBRIDGE}}     & \textbf{84.71 $\pm$ 0.0692} & \textbf{86.36 $\pm$ 0.0514} & \textbf{84.66 $\pm$ 0.0658} \\
\hline
\end{tabular}}
\vspace{-20pt}
\label{Table_1}
\end{center}
\end{table}
\vspace{-20pt}
\begin{table}[t]
\caption{Performance outcomes from ablation experiment [Unit: \%] (Mean $\pm$ Standard Deviation).}
\vspace{-15pt}
\begin{center}
\resizebox{\columnwidth}{!}{\begin{tabular}{|c|c|c|c|} 
\hline
\textbf{Method Variant} & \textbf{Accuracy} & \textbf{Sensitivity} & \textbf{Specificity} \\
\hline
w/o CBCL scores        & 82.13 $\pm$ 0.0882 & 77.73 $\pm$ 0.1281 & 82.27 $\pm$ 0.0798 \\
\hline
w/o Midpoint anchor       & 82.58 $\pm$ 0.0745 & 84.55 $\pm$ 0.0608 & 82.51 $\pm$ 0.0816 \\
\hline
w/o Trace-free deflation           & 83.84 $\pm$ 0.0712 & 85.91 $\pm$ 0.0549 & 83.78 $\pm$ 0.0794 \\
\hline
w/o Riemannian (in Euclidean)& 75.27 $\pm$ 0.1183 & 75.91 $\pm$ 0.1025 & 75.24 $\pm$ 0.1241 \\
\hline
w/o Gaussian RBF kernel $\Pi^{(\tau)}_{ij}$ & 83.79 $\pm$ 0.0697 & 85.45 $\pm$ 0.0562 & 83.74 $\pm$ 0.0739 \\
\hline
w/o Edge gate $g^{(\tau)}_{ij}$                     & 82.71 $\pm$ 0.0873 & 84.55 $\pm$ 0.0726 & 82.66 $\pm$ 0.0921 \\
\hline
w/o H-K tokens $S$            & 80.08 $\pm$ 0.0758 & 80.91 $\pm$ 0.0654 & 80.05 $\pm$ 0.0810 \\
\hline
w/o Cross-time attention  & 77.92 $\pm$ 0.1046 & 78.64 $\pm$ 0.0932 & 77.89 $\pm$ 0.1105 \\
\hline
w/o Koopman operator               & 79.69 $\pm$ 0.0989 & 80.91 $\pm$ 0.0887 & 79.65 $\pm$ 0.1034 \\
\hline
w/o  $\mathcal{L}_{\mathrm{dyn}}$     & 83.08 $\pm$ 0.0675 & 84.09 $\pm$ 0.0583 & 83.05 $\pm$ 0.0718 \\
\hline
w/o $\mathcal{L}_{\mathrm{proto}}$     & 79.52 $\pm$ 0.1985 & 71.36 $\pm$ 0.1291 & 79.78 $\pm$ 0.1612 \\
\hline
\textbf{Proposed}                  & \textbf{84.71 $\pm$ 0.0692} & \textbf{86.36 $\pm$ 0.0514} & \textbf{84.66 $\pm$ 0.0658} \\
\hline

\end{tabular}}
\vspace{-20pt}
\label{Table_2}
\end{center}
\end{table}

\vspace{5pt}
\subsection{Qualitative Evaluation}
\vspace{-2pt}
For each subject we exported \(\mathcal{E}^{(0)},\mathcal{E}^{(2)}\in\mathbb{R}^{N\times N}\) from the gate scores \(\tilde g_{ij}^{(\tau)}\), and computed a time-mixed edge map $\varpi  = \delta_0 \mathcal{E}^{(0)} + \delta_2 \mathcal{E}^{(2)} \in \mathbb{R}^{N\times N}$. We then averaged  $\varpi$ within class to obtain group means and retained the top 
3\% connections. 

\begin{figure}[htb]
  \centering
  \centerline{\includegraphics[width= 7 cm,height=3 cm]{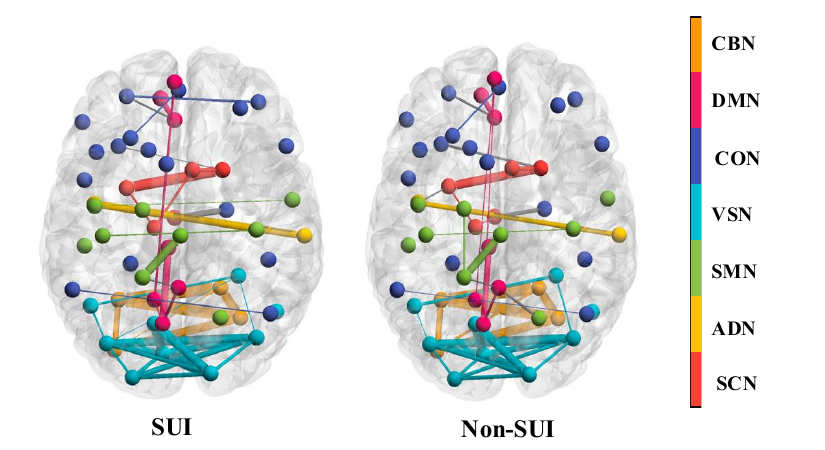}}
  \vspace{-5pt}
\caption{Axial view of group-mean, time-mixed edge weights (top $3\%$) for SUI and non-SUI cohorts across the seven networks: cerebellar (CBN), default mode (DMN), cognitive control (CON), visual (VSN), sensorimotor (SMN), auditory (ADN), and subcortical (SCN). Intra-network connections are color-coded by network; cross-network connections are shown in gray. Line width scales with edge weight, emphasizing relatively stronger connections.}
\vspace{-15pt}
\label{fig:fig2}
\end{figure}

As illustrated in Fig.~\ref{fig:fig2}, the non-SUI average depicted broadly distributed links from SMN and VSN toward CON and DMN, consistent with widely integrated coupling in normative development \cite{fair2009functional}. In SUI, high weight edges concentrated on CON--DMN coupling and CBN connections, with additional SCN links; in parallel, SMN/VSN connection weaken. This control--centric shift aligns with reports of altered fronto-parietal/DMN coordination and cerebellar involvement in adolescent risk phenotypes \cite{squeglia2009influence}. Overall, the differences in these map highlights strengthened CON--DMN/CBN pathways and reduced sensory–associative integration in SUI, a pattern coherent with prior neurodevelopmental findings \cite{fair2009functional,squeglia2009influence}.

\vspace{-10pt}
\section{Conclusions}
\vspace{-10pt}
We introduced \textit{NeuroBRIDGE}, a GNN based framework that aligns longitudinal FNCs in a Riemannian tangent space, fuses visits via dual self/cross attention, and models change with behavior-conditioned Koopman dynamics. On ABCD, it improved 4-year SUI prediction over SOTA baselines while yielding stable, geometry-aware representations and clinically interpretable network patterns. In future work, we plan to extend to multi-visit trajectories, incorporate additional modalities (diffusion/structural MRI and task fMRI), and explore alternative manifold-aware deep architectures for longitudinal modeling.



\bibliographystyle{IEEEbib}
\bibliography{refs}

@article{I_1,
  title={Braking and accelerating of the adolescent brain},
  author={Casey, Bety J and Jones, Rebecca M and Somerville, Leah H},
  journal={Journal of research on adolescence},
  volume={21},
  number={1},
  pages={21--33},
  year={2011},
  publisher={Wiley Online Library}
}

@article{fair2009functional,
  title={Functional brain networks develop from a “local to distributed” organization},
  author={Fair, Damien A and Cohen, Alexander L and Power, Jonathan D and Dosenbach, Nico UF and Church, Jessica A and Miezin, Francis M and Schlaggar, Bradley L and Petersen, Steven E},
  journal={PLoS computational biology},
  volume={5},
  number={5},
  pages={e1000381},
  year={2009},
  publisher={Public Library of Science San Francisco, USA}
}

@article{squeglia2009influence,
  title={The influence of substance use on adolescent brain development},
  author={Squeglia, Lindsay M and Jacobus, Joanna and Tapert, Susan F},
  journal={Clinical EEG and neuroscience},
  volume={40},
  number={1},
  pages={31--38},
  year={2009},
  publisher={SAGE Publications Sage CA: Los Angeles, CA}
}

@inproceedings{SPDNet,
  title={A riemannian network for spd matrix learning},
  author={Huang, Zhiwu and Van Gool, Luc},
  booktitle={Proceedings of the AAAI conference on artificial intelligence},
  volume={31},
  number={1},
  year={2017}
}

@article{BrainNetCNN,
  title={BrainNetCNN: Convolutional neural networks for brain networks; towards predicting neurodevelopment},
  author={Kawahara, Jeremy and Brown, Colin J and Miller, Steven P and Booth, Brian G and Chau, Vann and Grunau, Ruth E and Zwicker, Jill G and Hamarneh, Ghassan},
  journal={NeuroImage},
  volume={146},
  pages={1038--1049},
  year={2017},
  publisher={Elsevier}
}

@article{BNT,
  title={Brain network transformer},
  author={Kan, Xuan and Dai, Wei and Cui, Hejie and Zhang, Zilong and Guo, Ying and Yang, Carl},
  journal={Advances in Neural Information Processing Systems},
  volume={35},
  pages={25586--25599},
  year={2022}
}

@inproceedings{Evolvegcn,
  title={Evolvegcn: Evolving graph convolutional networks for dynamic graphs},
  author={Pareja, Aldo and Domeniconi, Giacomo and Chen, Jie and Ma, Tengfei and Suzumura, Toyotaro and Kanezashi, Hiroki and Kaler, Tim and Schardl, Tao and Leiserson, Charles},
  booktitle={Proceedings of the AAAI conference on artificial intelligence},
  volume={34},
  number={04},
  pages={5363--5370},
  year={2020}
}

@inproceedings{RGBM,
  title={Recurrent brain graph mapper for predicting time-dependent brain graph evaluation trajectory},
  author={Tekin, Alpay and Nebli, Ahmed and Rekik, Islem},
  booktitle={MICCAI Workshop on Domain Adaptation and Representation Transfer},
  pages={180--190},
  year={2021},
  organization={Springer}
}

@article{M_2,
  title={A Riemannian framework for tensor computing},
  author={Pennec, Xavier and Fillard, Pierre and Ayache, Nicholas},
  journal={International Journal of computer vision},
  volume={66},
  number={1},
  pages={41--66},
  year={2006},
  publisher={Springer}
}

@article{M_3,
  title={A differential geometric approach to the geometric mean of symmetric positive-definite matrices},
  author={Moakher, Maher},
  journal={SIAM journal on matrix analysis and applications},
  volume={26},
  number={3},
  pages={735--747},
  year={2005},
  publisher={SIAM}
}

@article{MM_0,
  title={Semi-supervised classification with graph convolutional networks},
  author={Kipf, TN},
  journal={arXiv preprint arXiv:1609.02907},
  year={2016}
}

@article{MM_2,
  title={Hamiltonian systems and transformation in Hilbert space},
  author={Koopman, Bernard O},
  journal={Proceedings of the National Academy of Sciences},
  volume={17},
  number={5},
  pages={315--318},
  year={1931}
}

@article{D_1,
  title={Baseline brain function in the preadolescents of the ABCD Study},
  author={Chaarani, B and Hahn, S and Allgaier, N and Adise, S and Owens, MM and Juliano, AC and Yuan, DK and Loso, H and Ivanciu, A and Albaugh, MD and others},
  journal={Nature neuroscience},
  volume={24},
  number={8},
  pages={1176--1186},
  year={2021},
  publisher={Nature Publishing Group US New York}
}

@article{D_2,
  title={NeuroMark: An automated and adaptive ICA based pipeline to identify reproducible fMRI markers of brain disorders},
  author={Du, Yuhui and Fu, Zening and Sui, Jing and Gao, Shuang and Xing, Ying and Lin, Dongdong and Salman, Mustafa and Abrol, Anees and Rahaman, Md Abdur and Chen, Jiayu and others},
  journal={NeuroImage: Clinical},
  volume={28},
  pages={102375},
  year={2020},
  publisher={Elsevier}
}

@inproceedings{D_3,
  title={Normalized dynamic time warping increases sensitivity in differentiating functional network connectivity in schizophrenia},
  author={Wiafe, Sir-Lord and Kinsey, Spencer and Iraji, Armin and Miller, Robyn and Calhoun, Vince D},
  booktitle={2025 47th Annual International Conference of the IEEE Engineering in Medicine and Biology Society (EMBC)},
  pages={1--4},
  year={2025},
  organization={IEEE}
}

@inproceedings{Diffusion_kernels,
  title={Diffusion kernels on graphs and other discrete structures},
  author={Kondor, Risi Imre and Lafferty, John},
  booktitle={Proceedings of the 19th international conference on machine learning},
  volume={2002},
  pages={315--322},
  year={2002}
}

@article{GRBF_kernel,
  title={Laplacian eigenmaps for dimensionality reduction and data representation},
  author={Belkin, Mikhail and Niyogi, Partha},
  journal={Neural computation},
  volume={15},
  number={6},
  pages={1373--1396},
  year={2003},
  publisher={MIT Press}
}

@article{prototype_loss,
  title={Prototypical contrastive learning of unsupervised representations},
  author={Li, Junnan and Zhou, Pan and Xiong, Caiming and Hoi, Steven CH},
  journal={arXiv preprint arXiv:2005.04966},
  year={2020}
}

@inproceedings{My_1,
  title={Genetics Encoded Joint Embedding of Multimodal Connectomes with Explainable Graph Neural Network for Schizophrenia Classification},
  author={Mazumder, Badhan and Wu, Lei and Calhoun, Vince D and Ye, Dong Hye},
  booktitle={2025 IEEE 22nd International Symposium on Biomedical Imaging (ISBI)},
  pages={1--5},
  year={2025},
  organization={IEEE}
}

@INPROCEEDINGS{My_2,
  author={Mazumder, Badhan and Wu, Lei and Calhoun, Vince D. and Hye Ye, Dong},
  booktitle={2025 47th Annual International Conference of the IEEE Engineering in Medicine and Biology Society (EMBC)}, 
  title={Unified Cross-Modal Attention-Mixer Based Structural-Functional Connectomics Fusion for Neuropsychiatric Disorder Diagnosis}, 
  year={2025},
  pages={1--6},
  organization={IEEE}
}

@INPROCEEDINGS{My_3,
  author={Mazumder, Badhan and Kotoski, Aline and Calhoun, Vince D. and Ye, Dong Hye},
  booktitle={2025 IEEE EMBS International Conference on Biomedical and Health Informatics (BHI)}, 
  title={NeuroKoop: Neural Koopman Fusion of Structural–Functional Connectomes for Identifying Prenatal Drug Exposure in Adolescents}, 
  year={2025},
  pages={1--7},
  organization={IEEE}
}

@article{My_4,
  title={KOCOBrain: Kuramoto-Guided Graph Network for Uncovering Structure-Function Coupling in Adolescent Prenatal Drug Exposure},
  author={Mazumder, Badhan and Wu, Lei and Wiafe, Sir-Lord and Calhoun, Vince D and Ye, Dong Hye},
  journal={arXiv preprint arXiv:2601.11018},
  year={2026}
}

\end{document}